\documentclass[letterpaper, paper,11pt]{AAS}		

\usepackage{bm}
\usepackage{amsmath}
\usepackage{subcaption}
\usepackage{multirow}
\usepackage[colorlinks=true, pdfstartview=FitV, linkcolor=black, citecolor= black, urlcolor= black]{hyperref}
\usepackage{overcite}
\usepackage{footnpag}			      	
\usepackage{gensymb}
\usepackage{tablefootnote}

\newcommand{\specialcell}[2][c]{%
  \begin{tabular}[#1]{@{}r@{}}#2\end{tabular}}

\PaperNumber{21-283}

\begin{document}

\title{REAL-TIME, FLIGHT-READY, NON-COOPERATIVE SPACECRAFT POSE ESTIMATION USING MONOCULAR IMAGERY}

\author{Kevin Black\thanks{Undergraduate Researcher, Texas Spacecraft Laboratory, The University of Texas at Austin}, Shrivu Shankar\footnotemark[1], Daniel Fonseka\footnotemark[1], Jacob Deutsch\footnotemark[1], Abhimanyu Dhir\footnotemark[1], and Maruthi R. Akella \thanks{Ashley H. Priddy Centennial Professorship in Engineering, Department of Aerospace Engineering and Engineering Mechanics, The University of Texas at
Austin, AAS Fellow, Email: makella@mail.utexas.edu}}

\maketitle{}

\begin{abstract}
A key requirement for autonomous on-orbit proximity operations is the estimation of a target spacecraft’s relative pose (position and orientation).  It is desirable to employ monocular cameras for this problem due to their low cost, weight, and power requirements. This work presents a novel convolutional neural network (CNN)-based monocular pose estimation system that achieves state-of-the-art accuracy with low computational demand. In combination with a Blender-based synthetic data generation scheme, the system demonstrates the ability to generalize from purely synthetic training data to real in-space imagery of the Northrop Grumman Enhanced Cygnus spacecraft. Additionally, the system achieves real-time performance on low-power flight-like hardware.
\end{abstract}

\section{Introduction}
Frequent proximity operations between spacecraft in orbit are a key requirement for current and future activities such as formation flying, debris removal\cite{RemoveDEBRIS}, and on-orbit servicing\cite{DARPAPhoenix,RestoreL}. Automation of these operations is crucial to their long-term value and viability. A fundamental requirement for autonomous proximity operations is the automatic detection of a target spacecraft’s relative position and attitude, often referred to as pose estimation. Oftentimes, as in the case of debris removal, the target is non-cooperative: it does not employ any active communications or special markings to assist the sensing spacecraft. Furthermore, many of these missions utilize only small spacecraft with strict power and mass requirements. Monocular cameras — which are small in size, ubiquitous in nature, and consume little power compared to other sensors such as stereo cameras or LIDAR —  are an ideal sensor for low-power non-cooperative pose estimation.

The problem of pose estimation from monocular imagery has been studied for many years. Early techniques focused on hand-engineered feature matching\cite{Cropp2001,Du2011}, but have been known to exhibit to poor robustness and generalization capabilities. This is particularly true in the space environment, which presents unique challenges such as low signal-to-noise ratio, harsh and varied lighting conditions, and a dynamic Earth background. However, over the past decade, nearly all computer vision tasks such as monocular pose estimation have become increasingly dominated by deep convolutional neural networks (CNNs). Compared to prior techniques, CNNs have been shown to be more resilient to noise and better able to generalize to previously unseen scenarios. It is no surprise that CNNs are now being applied for in-space monocular pose estimation. The 2019 Satellite Pose Estimation Challenge (SPEC)\cite{SatellitePoseEstimationChallenge}, hosted by Stanford University and the European Space Agency (ESA), saw all of its top-performing submissions employ CNN-based deep learning models. 

However, deep neural networks suffer from two major disadvantages: they require large amounts of labeled data to train and are computationally expensive to run. Due to the scarcity of labeled real images of spacecraft in orbit, most work on in-space monocular pose estimation has focused on synthetic or mockup-based imagery for both training and evaluation. For example, the dataset provided for the SPEC competition was the Spacecraft Pose Estimation Dataset (SPEED)\cite{Sharma2019,SPEED}, which consists of black-and-white synthetic images of the ESA Tango spacecraft as well as limited real images of a mockup. Models that perform well on a limited variety of artificial images will not necessarily transfer effectively to the real environment. There has also been relatively little discussion about the feasibility of running these models on flight-like hardware. Most CNNs require expensive computations that are optimized for a Graphics Processing Unit (GPU); however, the application of GPUs in the space domain is still a nascent field of study due to radiation exposure considerations\cite{GPU4S}. While traditional processors are still capable of running CNNs, many spacecraft, especially smaller satellites such as CubeSats, only have low-power processors that are incapable of running large, state-of-the-art CNN models in real-time.

This work presents a monocular pose estimation system that addresses many of the challenges described above. The core of the system is a novel CNN-based pose estimation architecture that is computationally inexpensive, yet maintains state-of-the-art accuracy. Similar to some prior works\cite{Park2019,UniAdelaide}, this architecture is a 3-step process consisting of an object detection CNN, a keypoint regression CNN, and an off-the-shelf perspective-n-point (PnP) solver. It also includes a final error prediction step which identifies bad pose estimates and replaces them with ``non-detections''. Furthermore, the entire architecture is trained using a novel synthetic data generation scheme that produces photorealistic images with many of the degradations present in real space imagery. Synthetic images are generated using Cycles, a physically-based, ray-tracing, production rendering engine packaged with the open-source 3D graphics software Blender\cite{Blender}.

Our aim is to present a nearly flight-ready system. As discussed above, real image performance, along with feasibility of using flight-like hardware, is essential for flight readiness. In contrast to most prior work, we analyze the ability of the pose estimation system to generalize to real images. These real images exhibit a wide range of poses, backgrounds, and the aforementioned degradations. In addition, we present a thorough benchmark of the system’s performance on an Intel Joule 570x single-board computer, the same type which flew on the NASA Johnson Space Center (JSC) Seeker CubeSat mission in September 2019\cite{Seeker1}.

We first provide a survey of related work. Then, we describe the details of our pose estimation architecture, as well as our synthetic data generation scheme using the Northrop Grumman Enhanced Cygnus vehicle as our target spacecraft. We next present results including the performance of our architecture on the SPEED dataset, the performance of our system on both synthetic and real images of Cygnus, and the hardware performance of our system on the Intel Joule 570x single-board computer.

\section{Related Work}
There is much pre-existing work in the context of spacecraft monocular pose estimation, including a recently published overview of methods in the field by Cassinis et al. (2019)\cite{Cassinis2019}. Early methods of pose estimation relied on classical, non-learning methods such as Hough line detection \cite{Cropp2001} and Canny edge detection \cite{Du2011} for producing raw image features. However, pivotal papers such as Krizhevsky et al. (2012)\cite{Krizhevsky2012} show that learned feature detection through the use of CNNs is both more accurate and robust for computer vision tasks.

In some cases, this has gone as far as solving for pose as the direct output of a CNN. Su et al. (2016) \cite{Su2016} show that by discretizing the viewpoint space, a CNN can be used to classify an input image directly into a binned pose. Kehlet et. al. (2017) \cite{Kehl2017} expand on this to add refinement and verification methods after the pose classification. Rather than modeling the problem as a classification task, Mahendran et al. (2017) \cite{Mahendran2017} show that directly regressing a quaternion representation of pose is a viable method where minimal, if any, post-processing is required. In this case, the task of pose estimation can be completely learned by an end-to-end CNN. Although more and more computer vision applications now rely on end-to-end CNNs, based on the results of the Satellite Pose Estimation Challenge \cite{SatellitePoseEstimationChallenge}, the best performing pose estimation models in the space domain still use classical pose solvers such as POSIT\cite{POSIT} or Perspective-n-Point (PnP)\cite{PNPRANSAC} attached to the output of a CNN-based keypoint detector. Much recent work \cite{Cassinis2019,Park2019,UniAdelaide} has shown such a keypoints-based pipeline to be one of the most viable methods for the specific task of monocular satellite pose estimation. Furthermore, all of these recent works first use an object detection step to crop to the region of interest in order to provide scale invariance to the keypoint model.

To compensate for uncertainties in keypoint predictions, Cassinis et al. (2020)\cite{Cassinis2020} show the problem can be modeled as predicting a heatmap (where peaks represent high-confidence locations of a given keypoint in the input image) for each keypoint as opposed to regressing coordinates directly. This provides some interpretability as each prediction can be visualized as a feature-heatmap when juxtaposed with the original image, and the 2D output allows for richer keypoint features to be fed to the accompanying pose solver. By instead encoding the heatmap as a pixel-level displacement field, pose estimation accuracy can further be improved as in Hou et. al.'s MobilePose\cite{MobilePose}.

While these architectural advances are shown to improve pose accuracy in aggregate, deployed pose applications also require additional guarantees on worst-case predictions. For spacecraft, Kalman filters are often applied to a stream of pose model predictions and sensor data to validate the likelihood of the estimated poses \cite{Kalman,Cassinis2019}. In CullNet\cite{CullNet}, Gupta et al. also show that bad pose estimates can be culled with the use of an additional error-predicting CNN. They use each candidate pose estimate to produce a corresponding 2D mask of the target object, then feed this mask along with the original image into a CNN which attempts to predict the accuracy of the pose estimate.

As discussed above, another common challenge of extra-terrestrial pose estimation is the scarcity of real data. This drives many methods (especially those that utilize CNNs) to employ transfer learning \cite{Cassinis2019}. In some cases, this requires initially training the keypoint model on a completely separate domain to learn key CNN filters that are still generalizable to the target domain of space \cite{Sha2018}. Other work attempts to use visual simulations to generate artificial training data. Notably, this includes the OpenGL-rendered SPEED dataset \cite{SPEED,Sharma2019}, as well as the Unreal Engine-based simulator presented in Proenca and Gao (2019)\cite{Proenca2019}. Generative Adversarial Networks (GANs) have also been employed in similar contexts to improve the realism of artificial training data \cite{DeepURL}.

\section{Methods}
The proposed pose estimation architecture is composed of a 3-step process. An object detection network first determines the 2D bounding box of the target spacecraft, and then feeds a cropped region of interest (RoI) to a separate keypoint regression network. The keypoint regression network regresses the 2D locations of predetermined 3D surface keypoints on the spacecraft model, from which the full pose is obtained using an off-the-shelf Perspective-n-Point (PnP) solver. This overall architecture was chosen because prior works\cite{SatellitePoseEstimationChallenge,Cassinis2019,Park2019,UniAdelaide}, as well as our own experiments, have found that such 3-step processes tend to outperform other techniques such as direct pose regression or classification. The final error prediction step predicts the approximate error of a given pose estimate. These error predictions can then be used to reject bad pose estimates and produce ``non-detections'' instead.

\subsection{Keypoint Selection}
We utilize surface keypoints for pose estimation rather than the corners of the 3D bounding box. The bounding box corners are often used in systems that need to handle multiple objects; however, as we only need to consider a single known object, we use surface keypoints due to their stronger correspondence with local image features.

To select keypoints, we exploit our access to the 3D mesh model of the Cygnus spacecraft that is used for synthetic image generation. A user-specified number of keypoints ($n$) are generated randomly using a sample elimination algorithm\cite{SampleElimination} that produces Poisson disk sample sets. The result is that the generated points are approximately ``evenly spread out'' in 3D space, serving well for the task of pose estimation for any given number of points. We find that using $n = 20$ keypoints produces good results, and use that value for all experiments.

For evaluation on the SPEED dataset, we use the same 11 surface keypoints as in Park et. al. (2019)\cite{Park2019} and Chen et. al. (2019)\cite{UniAdelaide}.

\subsection{Object Detection}
For the object detection network, we use off-the-shelf models from the TensorFlow Object Detection API \cite{ObjectDetection}. These are easy to train and include access to weights pretrained on the COCO 2017 dataset \cite{COCO}. We select the SSD MobileNetv2 architecture\cite{MobileNetv2} with an input resolution of $320 \times 320$, as it provides adequate accuracy while being the least computationally expensive architecture with an available pretrained checkpoint.

We operate in the context of a guaranteed single spacecraft in every image, so we always take the top bounding box detection regardless of confidence. To produce the final RoI, the detected bounding box is made square by setting the length of its shorter side to that of its longer side and then expanded by 25\% to ensure that the entire spacecraft lies inside.

\subsection{Keypoint Regression}
\begin{figure}[ht]
	\centering\includegraphics[width=\linewidth]{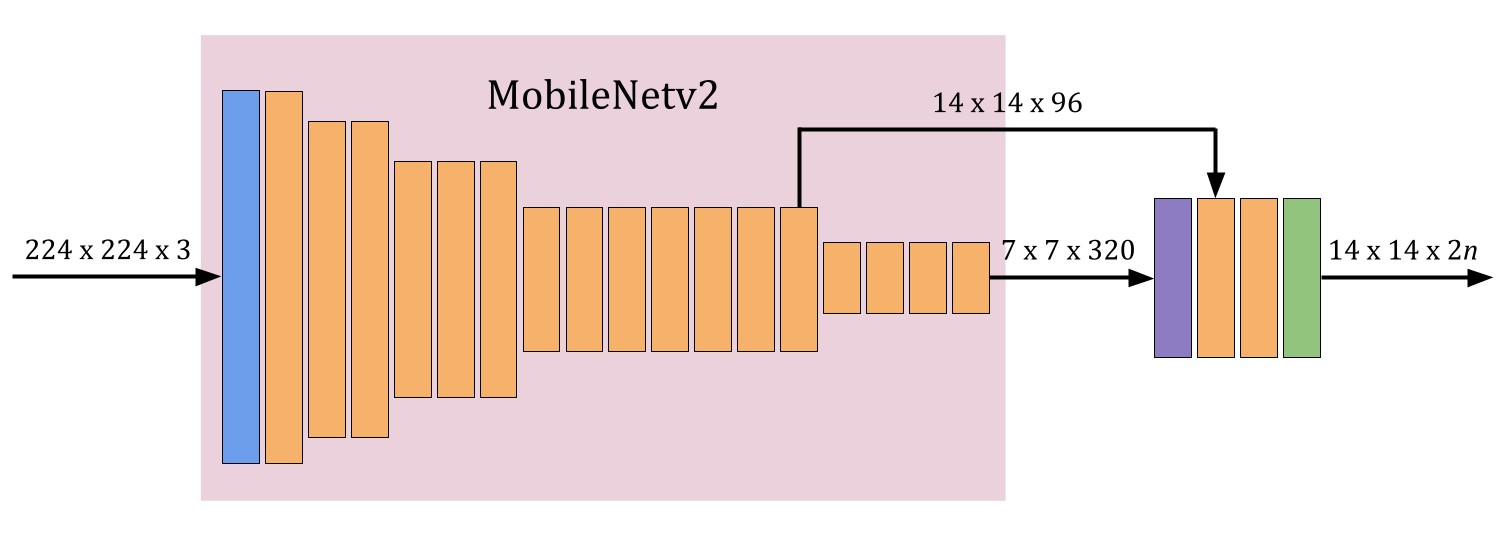}
	\caption{Layer diagram for the keypoint regression network. Blue rectangles are convolutional layers, purple are transposed convolutional layers, orange are inverted residual blocks, and green is the output $1 \times 1$ convolution.}
	\label{fig:network_diagram}
\end{figure}

\begin{figure}[htp]
    \centering
    \begin{subfigure}[t]{0.45\textwidth}
      \includegraphics[width=\linewidth]{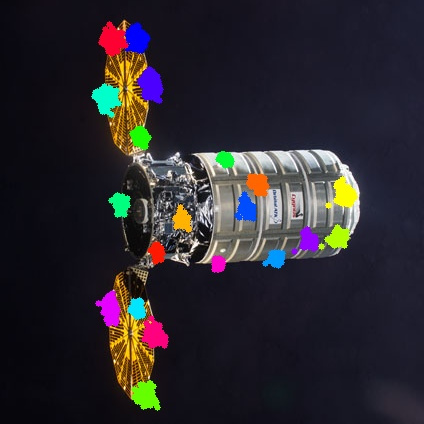}
    \end{subfigure}\hfil 
    \begin{subfigure}[t]{0.45\textwidth}
      \includegraphics[width=\linewidth]{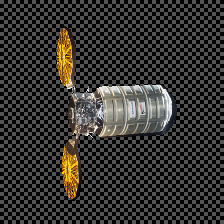}
    \end{subfigure}\hfil 
	\caption{Left: an example output of the keypoint regression network. Each same-colored cluster of points is the 196 predictions for each keypoint. Right: an example of a mask cutout fed to the error prediction network. The background has been filled with a checkered pattern for illustration purposes; however, in practice those pixels would all be set to black.}
	\label{fig:keypoint_example}
\end{figure}

After bounding box detection, the image is cropped to the RoI and rescaled to a resolution of $224\times224$. The resulting image is then fed to the keypoint regression network.

The keypoint regression network (Figure~\ref{fig:network_diagram}) is based on a MobileNetv2\cite{MobileNetv2} backbone and follows the general architecture of MobilePose\cite{MobilePose}. Unlike MobilePose, we keep the original MobileNetv2 network and cut it off before its final convolutional layer, producing a $7 \times 7 \times 320$ feature map. The feature map is upsampled back to $14 \times 14$ with a transposed convolutional layer, followed by a skip connection via concatenation to the last layer in MobileNetv2 with the same $14 \times 14$ scale. This is followed by two more of MobileNetv2's inverted residual blocks, and then a final $1 \times 1$ convolution to produce an output tensor with dimension $14 \times 14 \times 2n$, where $n$ is the number of keypoints.

Each of the $14 \times 14$ entries in the output tensor make a prediction for the 2D locations of all $n$ keypoints, meaning each keypoint has 196 predictions. As in MobilePose, the network does not predict the keypoints' 2D locations directly, but rather each of the $14 \times 14$ output locations predict an offset from their corresponding location in the image to each of the keypoint's 2D locations. The network is trained with L1 (mean absolute error) loss as to be more lenient to outliers, since not every output location may be able to accurately predict the location of every keypoint.

The goal of the downsampling-upsampling architecture is to combine both global and local image features to obtain more accurate keypoint predictions. As with the object detection network, the depthwise separable convolutions of MobileNetv2 drastically reduce the number of computations compared to typical CNNs and allow real-time inference on low-power devices.

\subsection{Pose Estimation}
The keypoint regression network produces $14 \times 14 = 196$ predictions for each of $n$ keypoints, resulting in $196n$ total correspondences between 3D keypoints and their 2D locations in the image. The spacecraft pose can then be obtained using off-the-shelf PnP solver software. However, the keypoint regression is deliberately designed to be lenient to outliers, so random sample consensus (RANSAC)\cite{PNPRANSAC} is also necessary to make the solution more robust. We use the \textit{solvePnPRansac} function from the open-source OpenCV \cite{OpenCV} library configured to use the EPnP \cite{EPnP} algorithm, which was found to provide the best balance of speed and accuracy.

\subsection{Error Prediction}

Similar to CullNet\cite{CullNet}, a 2D binary mask of the target spacecraft is first created using the estimated pose. This is done quickly using OpenGL. However, rather than concatenating the mask with the original image, the mask is instead used to ``cut out'' the shape  of the target spacecraft; i.e., inside the mask, the original image is kept and outside the mask, every pixel is set to zero. All of these operations occur within the $224 \times 224$ cropped RoI. The resulting cutout is used as the input to the error prediction network.


Rather than trying to predict an error derived from the final pose solution, we found it more accurate to instead predict an error derived from the output of the keypoint regression network. The error metric we use is the mean Euclidean distance between all 2D keypoint estimates and their corresponding ground truth locations:
\begin{equation}
    E_k = \frac{1}{196n} \sum_{i=0}^{n}{\sum_{j=0}^{196}{||\bm{k_{i} - \bm{\hat{k}_{i, j}}}||}}
\end{equation}
where $\bm{k_i}$ is the ground truth location of the $i$th keypoint, and $\bm{\hat{k}_{i,j}}$ is the $j$th estimate for the $i$th keypoint from the keypoint regression network. Keypoint locations are expressed as 2D pixel coordinates within the $224 \times 224$ cropped RoI. The task of the error prediction network is thus to take a cutout as the input and regress $E_k$.

The cutout method allows the error prediction network to reuse nearly the full architecture and weights of the keypoint regression network, with only the final $1 \times 1$ convolution replaced with a fully connected layer. We found the ability to initialize the error prediction network with the weights of the keypoint regression network to significantly aid training.

Once the network produces a prediction for the keypoint error, $\hat{E_k}$, a threshold must be chosen above which pose estimates are rejected. Using the synthetic test set, we choose 20 pixels as a good balance between rejecting bad estimates and keeping good ones.

\subsection{Synthetic Data Generation}
\begin{figure}[htb]
    \centering 
\begin{subfigure}[t]{0.25\textwidth}
  \includegraphics[width=\linewidth]{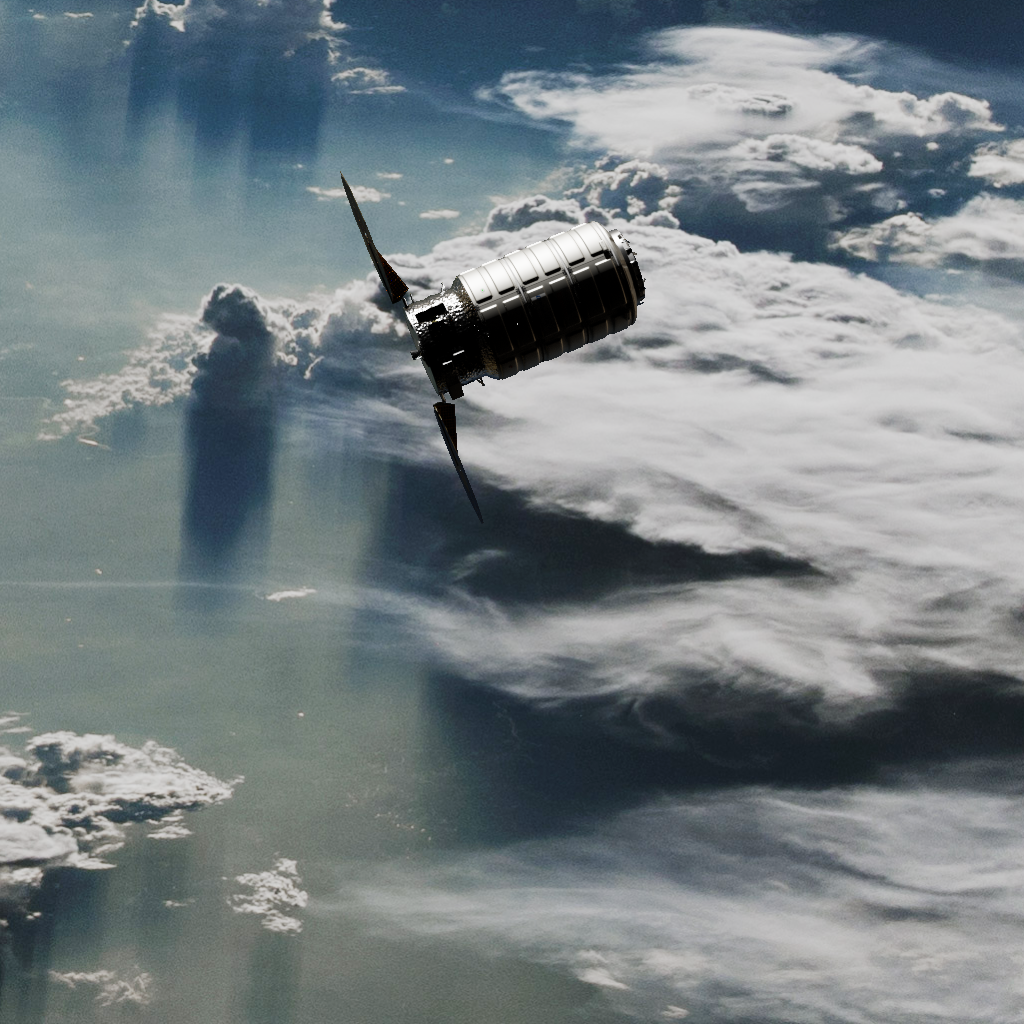}
  \caption{Cygnus with no augmentations in front of a real Earth background}
\end{subfigure}\hfil 
\begin{subfigure}[t]{0.25\textwidth}
  \includegraphics[width=\linewidth]{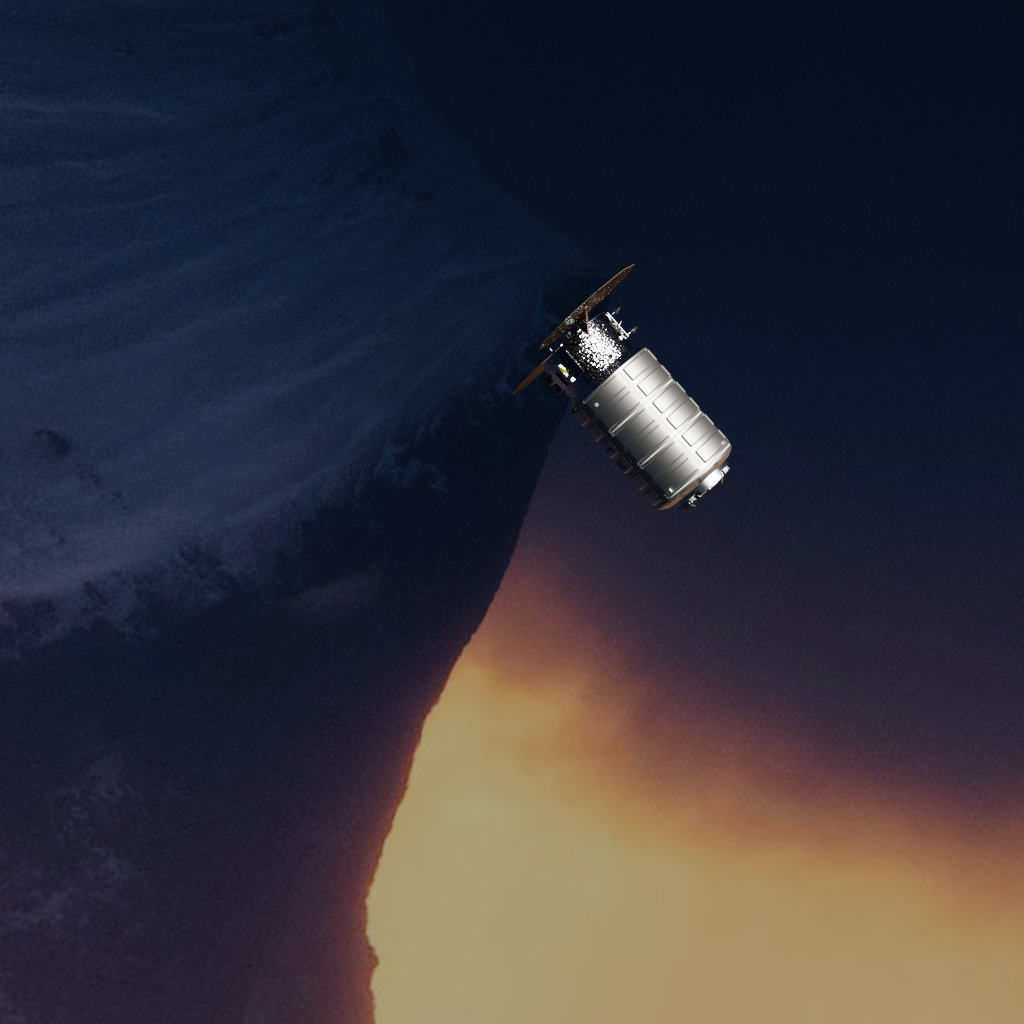}
  \caption{Cygnus in front of a randomized background}
\end{subfigure}\hfil 
\begin{subfigure}[t]{0.25\textwidth}
  \includegraphics[width=\linewidth]{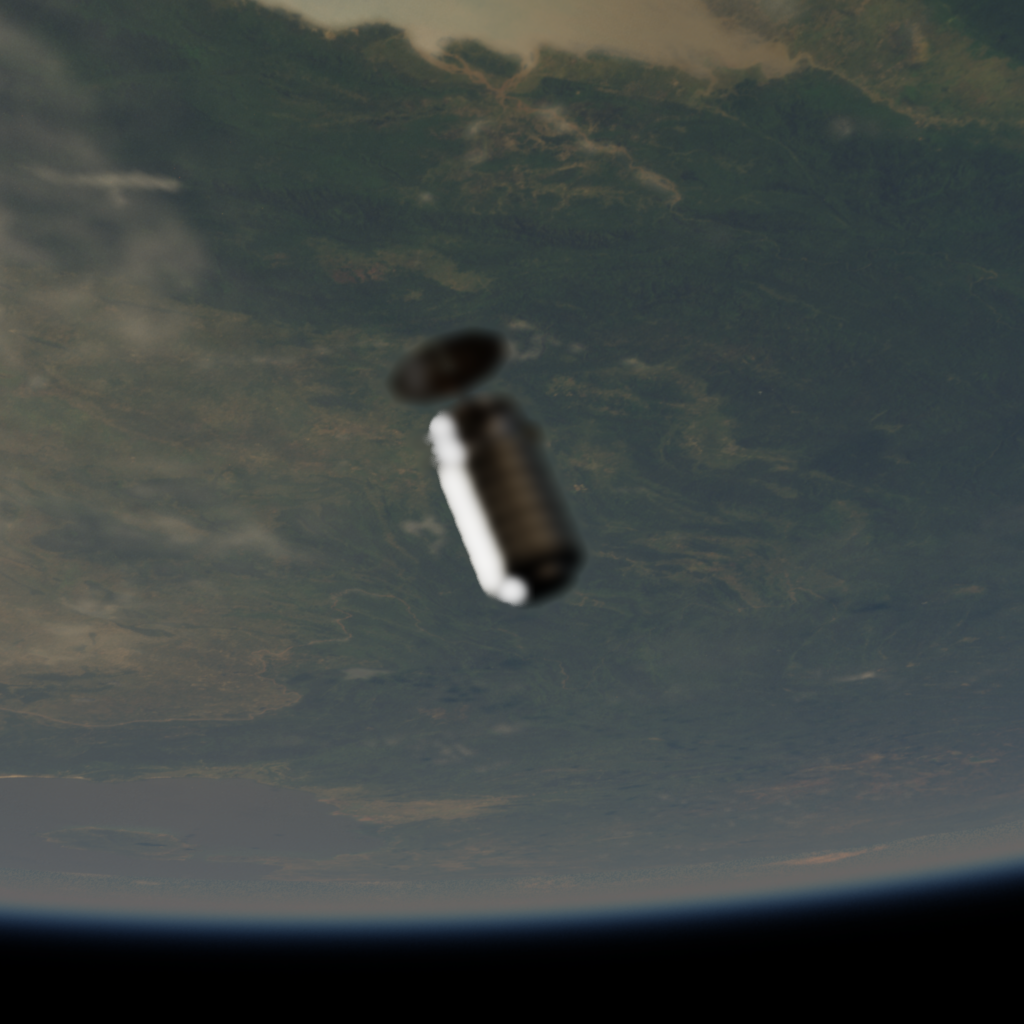}
  \caption{Cygnus with a blur effect}
\end{subfigure}

\medskip
\begin{subfigure}{0.25\textwidth}
  \includegraphics[width=\linewidth]{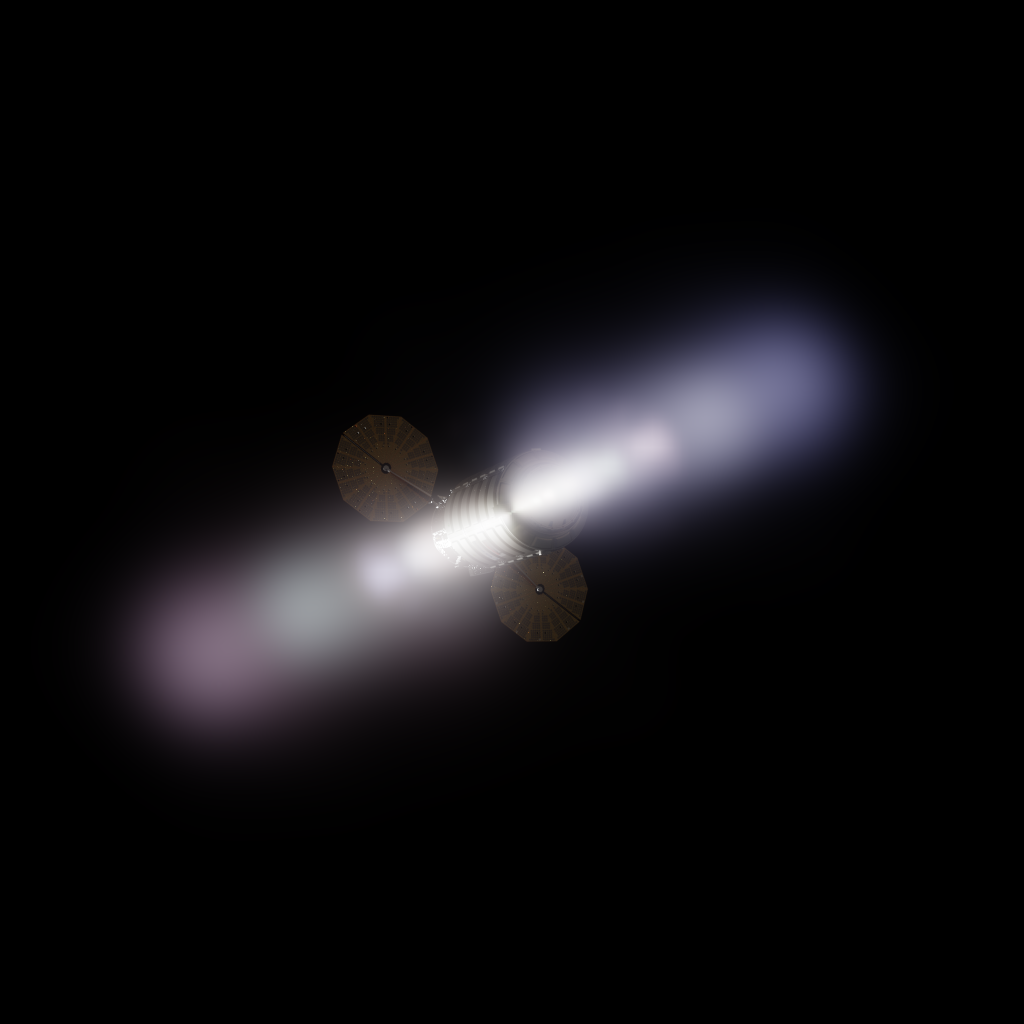}
  \caption{Cygnus with a lens flare effect}
\end{subfigure}\hfil 
\begin{subfigure}{0.25\textwidth}
  \includegraphics[width=\linewidth]{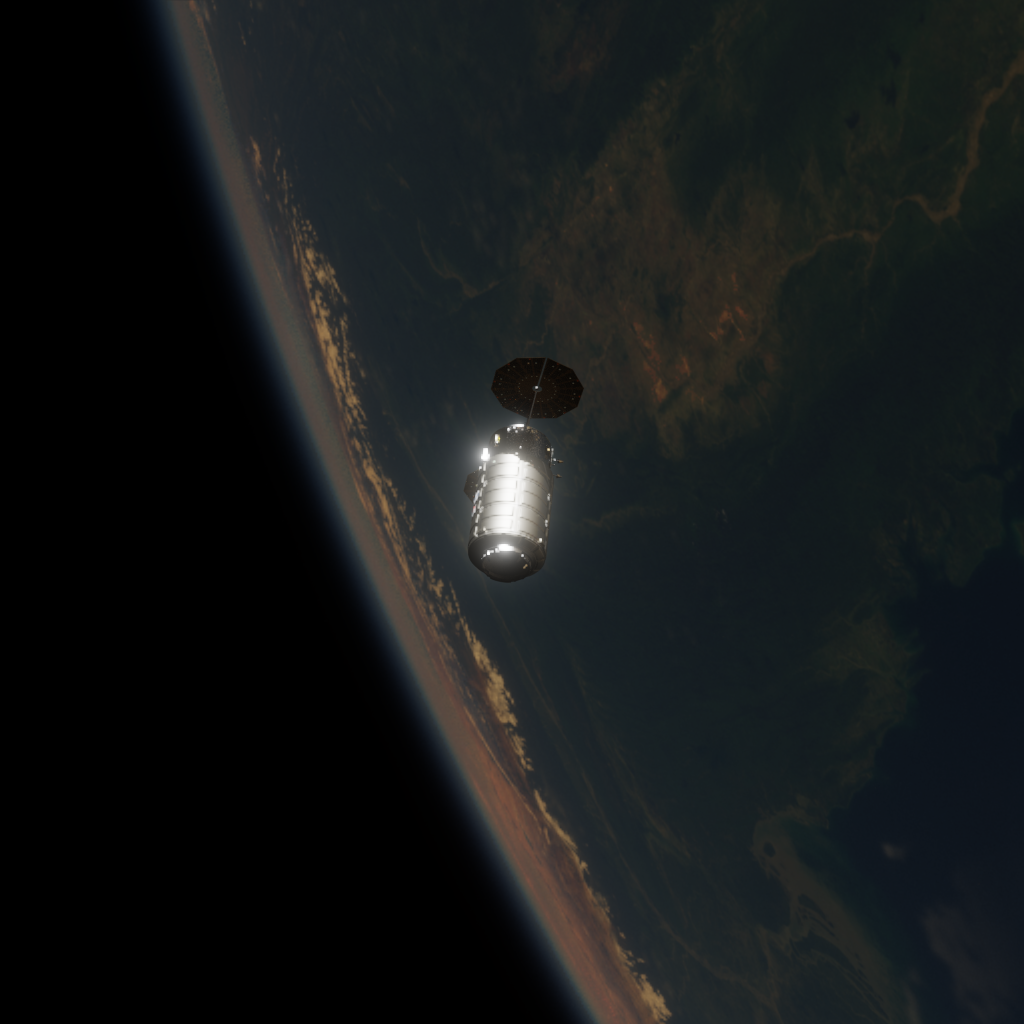}
  \caption{Cygnus with a ``fog glow'' effect}
\end{subfigure}\hfil 
\begin{subfigure}{0.25\textwidth}
  \includegraphics[width=\linewidth]{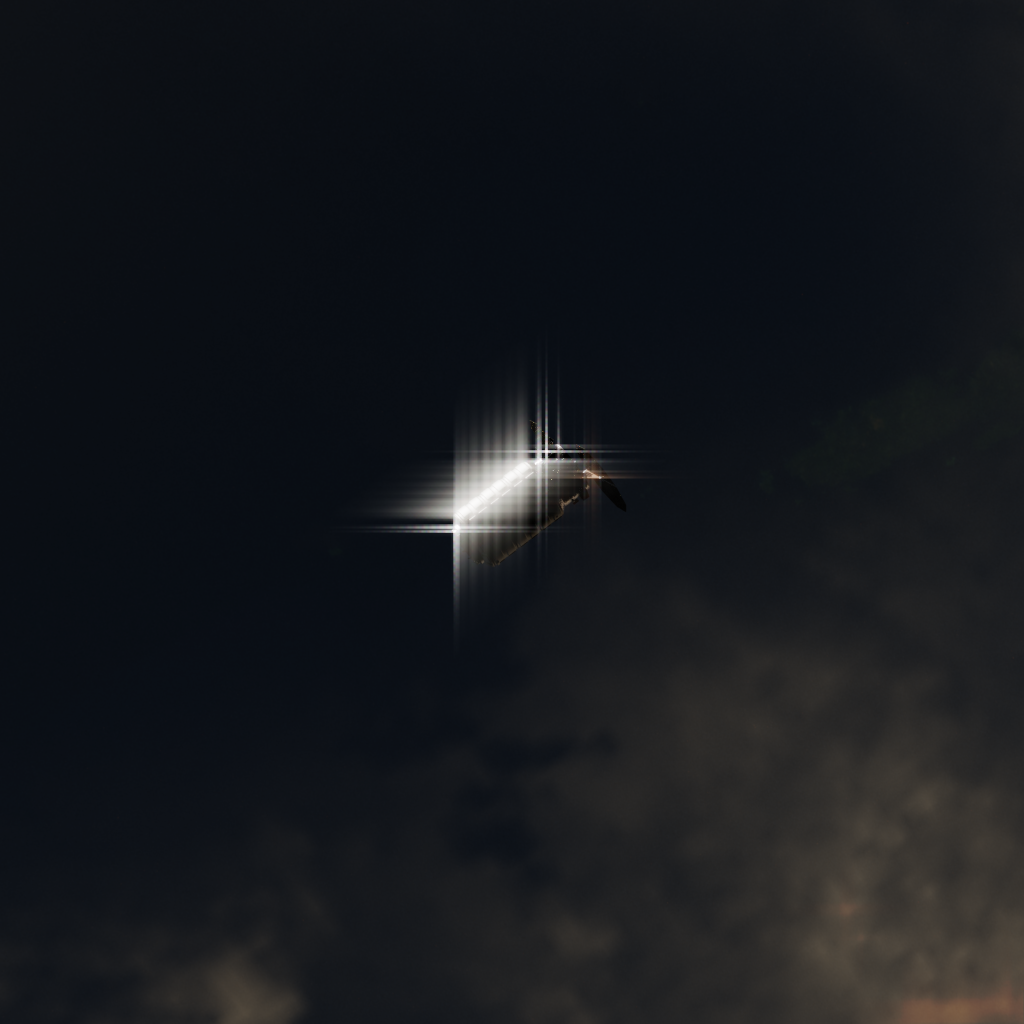}
  \caption{Cygnus with a ``simple star'' effect}
\end{subfigure}
\caption{Example synthetic images with and without augmentations. The bottom row all falls under the category of ``glare and lens flares''.}
\label{fig:synthetic_examples}
\end{figure}

Due to the scarcity of labeled real images, we generate synthetic images for our experiments with the Northrop Grumman Enhanced Cygnus spacecraft. Images are generated using Blender \cite{Blender}, an open-source 3D modeling suite, and its Cycles\footnote{\url{https://www.cycles-renderer.org/}} rendering engine. Cycles is a physically-based, ray-tracing, production-quality rendering engine. In comparison to prior work which has used OpenGL shaders\cite{Sharma2019} or Unreal Engine\cite{Proenca2019}, we use Cycles to better mimic real camera parameters and produce more photorealistic images. We obtain our model of the Cygnus spacecraft from an open-source GitHub repository\footnote{\url{https://github.com/brickmack/Blender-Spaceflight-Models}}.

Blender's powerful Python API also allows easy configuration of the camera position, lighting angle, and spacecraft pose. We use a custom-built library\footnote{\url{https://autognc-starfish.readthedocs.io}} to leverage this API and enable flexible, automated generation of labeled synthetic images.

To better simulate images taken in the true space environment, we additionally use Blender to add custom augmentations to the synthetic data. These augmentations include various types of randomized glare, lens flares, blur, and background images. Two types of background images are used. The first is real satellite photos of Earth, intended to further improve realism. The second is completely randomized non-realistic background images, intended to reduce overfitting during training and improve generalization to the variance present in real images. These augmentations to the training data have led to a marked increase in the performance of the pose estimation system on real images.

\subsection{Training Details}

The object detection network is initialized with weights pre-trained on the COCO 2017 dataset\cite{COCO}. It is trained for 40,000 steps using the Adam optimizer with a batch size of 10 and an initial learning rate of 0.001 that decays to 0.0001 at 20,000 steps and 0.00001 at 30,000 steps. Random flips, 90\degree~rotations, and crops are applied.

The MobileNetv2 portion of the keypoint regression network is initialized with weights pre-trained on the ImageNet dataset\cite{ImageNet}. The entire network is then trained for 300 epochs using the Adam optimizer with a batch size of 60 and an initial learning rate of 0.001 that decays to 0.0001 at 100 epochs and begins decaying exponentially after 200 epochs as $0.001 e^{-0.05(\text{epoch} - 200)}$. The RoIs used for training are derived from the ground-truth bounding box labels with random translations and expansions applied. Random flips, 90\degree~rotations, brightness, contrast, and saturation augmentations are also applied.

The error prediction network is initialized with weights from the fully trained keypoint regression network. The training data is a holdout test set consisting of synthetic images never seen by the object detection or keypoint regression networks. The network is trained for 50 epochs using the Adam optimizer with a batch size of 64 and a learning rate of 0.001.

\section{Results}
\subsection{Evaluation Metrics}
For evaluating the object detection network individually, we use the standard intersection-over-union (IoU) of the detected and ground-truth bounding boxes. We also report an additional metric we call RoI accuracy, which is the proportion of images in which the final region of interest (the detected bounding box squared and expanded by 25\%) contains the entire ground-truth bounding box.

Pose is represented as a pair $(\bm{q, t})$ where $\bm{q}$ is a rotation represented as a quaternion and $\bm{t}$ is a 3-dimensional translation vector. Applied together, $(\bm{q, t})$ align the camera reference frame with the target reference frame. 4 metrics will be used to measure the error between a ground-truth pose $(\bm{q, t})$ and predicted pose $(\bm{\hat{q}, \hat{t}})$. The first metric measures the rotation error, computed as:
\begin{equation}
    E_R = 2 \arccos{|\bm{q} \cdot \bm{\hat{q}}|}
\end{equation}
where $\cdot$ is the vector dot product. $E_R$ corresponds to the angle of the smallest rotation that aligns $\bm{\hat{q}}$ and $\bm{q}$. Another two metrics measure the translation error:
\begin{equation}
    E_T = ||\bm{t} - \bm{\hat{t}}||
\end{equation}
\begin{equation}
    E_{TN} = \frac{||\bm{t} - \bm{\hat{t}}||}{||\bm{t}||}
\end{equation}
$E_T$ is the distance between the ground truth and predicted translations in a real unit, such as meters; $E_{TN}$ is this distance normalized by the ground truth distance between the target and the camera. A final metric combines the two errors:
\begin{equation}
    E_C = E_R + E_{TN}
\end{equation}

For experiments with Cygnus, these pose metrics will be reported twice: once for the full dataset, and once with rejected pose estimates removed using the error prediction network. The proportion of pose estimates that were rejected will also be reported.

\subsection{SPEED Dataset}
\begin{figure}[htb]
    \centering 
\begin{subfigure}[t]{0.45\textwidth}
  \includegraphics[width=\linewidth]{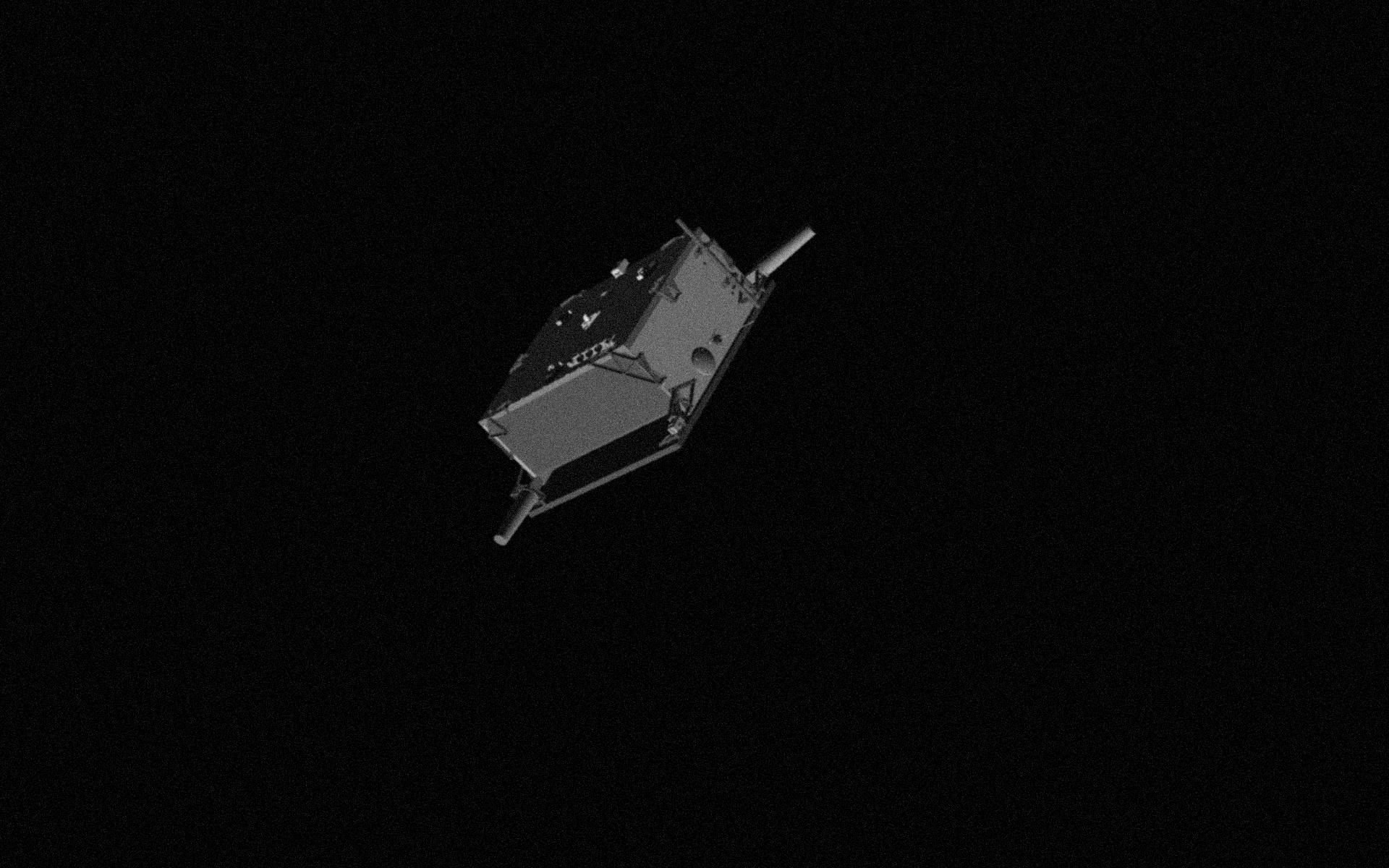}
\end{subfigure}\hfil 
\begin{subfigure}[t]{0.45\textwidth}
  \includegraphics[width=\linewidth]{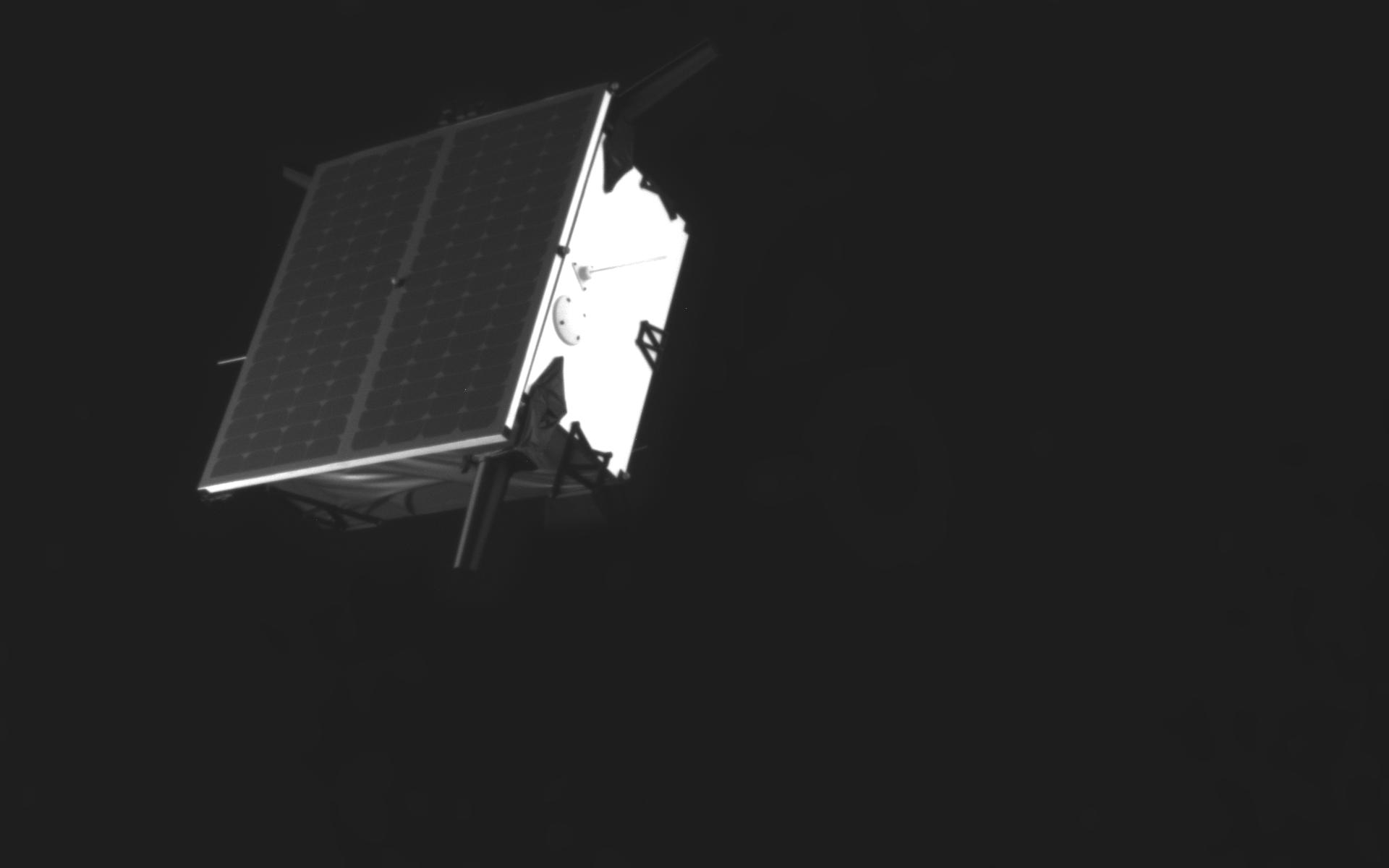}
\end{subfigure}\hfil 
\caption{A synthetic image (left) and real image of a mockup (right) from the SPEED dataset}
\label{fig:speed_images}
\end{figure}

First, our pose estimation system is evaluated on the SPEED dataset\cite{SPEED} in order to provide a comparison to the top submissions in the Satellite Pose Estimation Challenge (SPEC)\cite{SatellitePoseEstimationChallenge}. The SPEED dataset consists of 12,000 training images and 2,998 test images of the ESA Tango spacecraft, all grayscale and synthetic. 300 real grayscale images of a Tango mockup are also included. The images have a resolution of $1900 \times 1200$ and a horizontal field of view of 35.1\degree. The distance of Tango from the camera ranges from 3 to 40.5 meters, with bounding box side length ranging from 27 pixels to 1432 pixels with a mean of 371 pixels.

\begin{table}[htbp]
	\fontsize{10}{10}\selectfont
    \caption{Satellite Pose Estimation Challenge Comparison}
    \label{tab:speed}
        \centering 
   \begin{tabular}{r | l | c | c | c}
      \hline 
      SPEC Rank & Name & Mean $E_C$ (synthetic) & Mean $E_C$ (real) & \# parameters (millions)\\
      \hline 
      1      & \texttt{UniAdelaide}\cite{UniAdelaide} & 0.0094 & 0.3752 & 60+\\
      2      & \texttt{EPFL\_cvlab}\footnotemark & 0.0215 & 0.1140 & 60+ \\
             & \textbf{Ours} & \textbf{0.0409} & \textbf{0.2918} & \textbf{6.9} \\
      3      & \texttt{pedro\_fairspace}\cite{Proenca2019} & 0.0570 & 0.1555 & 60+ \\
      4      & \texttt{stanford\_slab}\cite{Park2019} & 0.0626 & 0.3951 & 11.2 \\
   \end{tabular}
\end{table}

\footnotetext{\url{https://indico.esa.int/event/319/attachments/3561/4754/pose_gerard_segmentation.pdf}}

Table~\ref{tab:speed} shows our system ranked alongside the top 4 submissions to the SPEC competition. The labels for the SPEED test set have not been released publicly, so we are only able to report the $E_C$ scores obtained from the SPEC post-mortem scoring server\footnote{\url{https://kelvins.esa.int/pose-estimation-challenge-post-mortem/}}. We are also not able to test the error prediction step, since a pose estimate must be returned for every image. The competition ranks submissions based on the synthetic mean $E_C$ score only. The parameter counts are obtained from the publications corresponding to each of the submissions, and roughly correspond to a network's computational needs.

Our pose estimation system achieves competitive accuracy, placing 3rd in both synthetic and real image score while having the fewest parameters. The only other top submission with a comparable network size is that of \texttt{stanford\_slab}.

\newpage

\subsection{Cygnus: Synthetic Data}
\begin{table}[htbp]
	\fontsize{10}{10}\selectfont
    \caption{Dataset Breakdown}
   \label{tab:dataset_breakdown}
        \centering 
   \begin{tabular}{l | l}
      \hline 
      Description & Number of Images\\
      \hline 
      No augmentations & 3,000\\
      Glare and lens flares & 3,000\\
      Blur & 3,000\\
      Spacecraft partially out-of-frame & 3,000\\
      No augmentations, real Earth background & 2,000\\
      Glare and lens flares, real Earth background & 2,000\\
      Blur, real Earth background & 2,000\\
      No augmentations, randomized background & 2,000\\
      \textbf{Total} & \textbf{20,000}
   \end{tabular}
\end{table}

Table~\ref{tab:dataset_breakdown} shows the composition of the synthetic dataset we use for all experiments with the Cygnus spacecraft. All images have resolution $1024 \times 1024$ with a field of view of 39.6\degree. Position, orientation, and lighting angle are all randomized. Target distance from the camera is randomized from 35 to 75 meters. Not accounting for images with Cygnus partially out of frame, bounding box side length ranges from 76 to 459 pixels, with a mean of 232 pixels. The dataset is split into 64\% training, 16\% validation, and a 20\% holdout test images.

\begin{table}[htbp]
	\fontsize{10}{10}\selectfont
    \caption{Synthetic Dataset Performance}
   \label{tab:synthetic_results}
        \centering 
   \begin{tabular}{r | l | l | l}
      \cline{2-4} 
      & Metric & Median & Mean\\
      \hline 
      \multirow{2}{*}{Object Detection Metrics}
      & RoI Accuracy & - & 0.98\\
      & IoU & 0.95 & 0.92\\
      \hline
      \multirow{4}{*}{Pose Metrics}
      & $E_R$ (deg) & 3.22 & 25.96\\
      & $E_T$ (meters) & 0.93 & 85.49\\
      & $E_{TN}$ & 0.017 & 1.565\\
      & $E_C$ & 0.074 & 2.018\\
      \hline
      \multirow{5}{*}{\specialcell{Rejected Estimates \\ ($\hat{E_k} > 20$) Removed}}
      & Proportion Rejected & - & 0.18\\
      & $E_R$ (deg) & 2.64 & 6.45\\
      & $E_T$ (meters) & 0.71 & 1.08\\
      & $E_{TN}$ & 0.013 & 0.019\\
      & $E_C$ & 0.062 & 0.132\\
   \end{tabular}
\end{table}

Table~\ref{tab:synthetic_results} shows the performance of the pose estimation system on the aforementioned holdout set. The object detection network performs very well: the resulting RoI includes the entire spacecraft in 98\% of images. The pose errors, on the other hand, are characterized by extreme outliers: due to the various augmentations, the system produces a few ``bad estimates'' on the most difficult images. (This includes, of course, the 2\% of images where the initial RoI is inaccurate.) In particular, the translation errors are unbounded, and so a few extreme outliers can severely inflate the mean. However, the error prediction network easily detects these extreme outliers. As mentioned before, the $\hat{E_k}$ threshold is chosen using this synthetic dataset to provide a good balance between rejecting bad estimates and keeping good ones. The chosen threshold of 20 pixels results in 18\% of estimates being rejected while bringing down the mean errors to reasonable values.

Overall, as with the SPEED dataset, it is expected that the system will perform well on the same type of synthetic data it is trained on. The synthetic image error provides a rough lower bound on the achievable real image error.

\subsection{Cygnus: Real Data}
\begin{figure}[h]
    \centering 
\begin{subfigure}[t]{0.3\textwidth}
  \includegraphics[width=\linewidth]{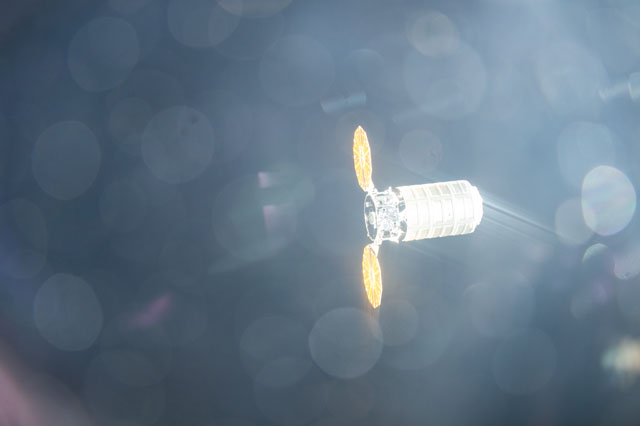}
\end{subfigure}\hfil 
\begin{subfigure}[t]{0.3\textwidth}
  \includegraphics[width=\linewidth]{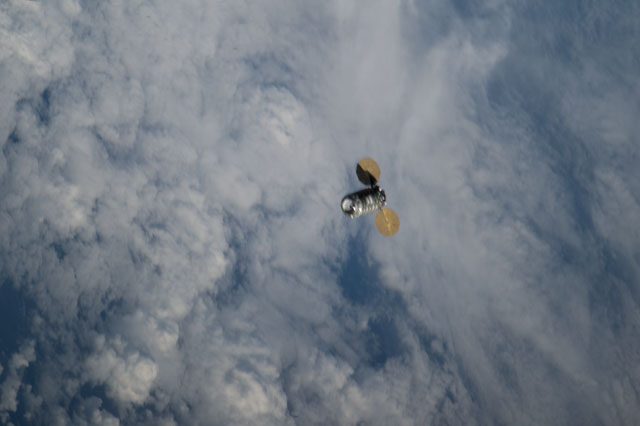}
\end{subfigure}\hfil 
\begin{subfigure}[t]{0.3\textwidth}
  \includegraphics[width=\linewidth]{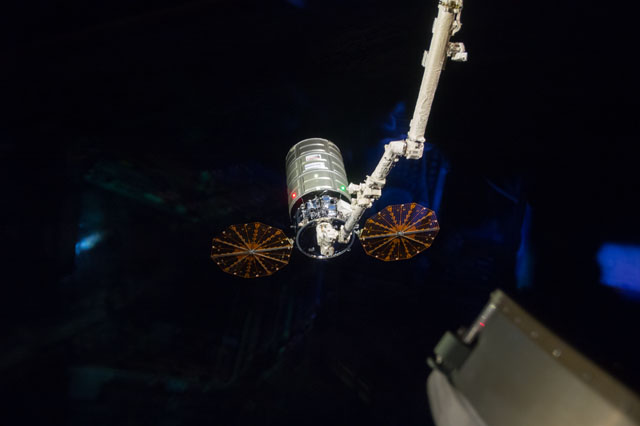}
\end{subfigure}

\medskip
\begin{subfigure}[t]{0.3\textwidth}
  \includegraphics[width=\linewidth]{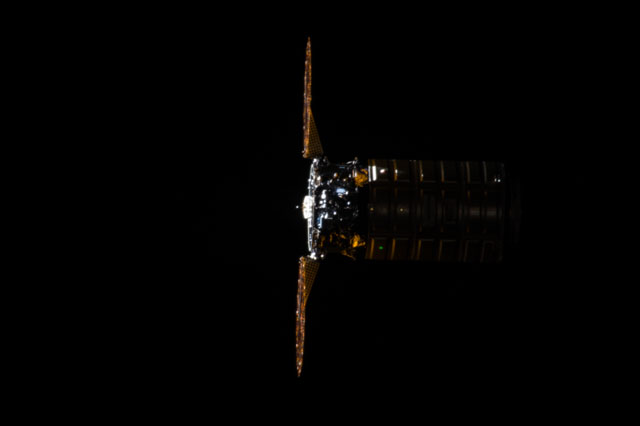}
\end{subfigure}\hfil 
\begin{subfigure}[t]{0.3\textwidth}
  \includegraphics[width=\linewidth]{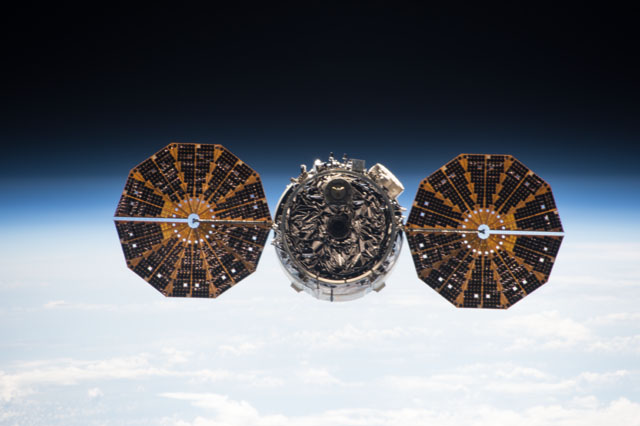}
\end{subfigure}\hfil 
\begin{subfigure}[t]{0.3\textwidth}
  \includegraphics[width=\linewidth]{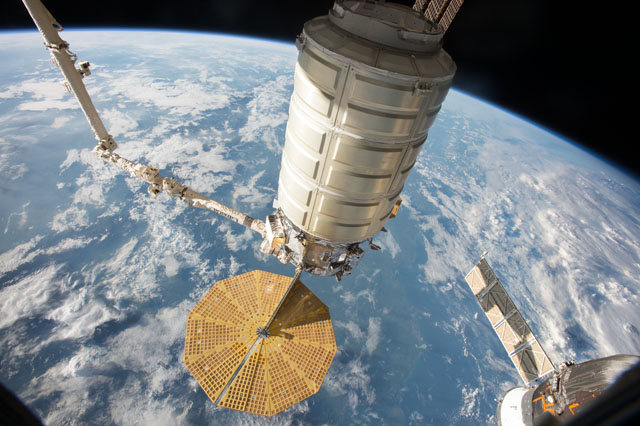}
\end{subfigure}

\caption{Examples of real images}
\label{fig:real_images}
\end{figure}
The real dataset consists of 540 photos of the Cygnus spacecraft taken in orbit. The photos are obtained from NASA\footnote{\url{https://images.nasa.gov/}}. They cover a variety of poses and lighting conditions, and include degradations such as glare and motion blur. The field of view varies from 1.8\degree~to 86.9\degree, and bounding box side length varies from 54 pixels to 744 pixels with a mean of 231 pixels.

\begin{table}[htbp]
	\fontsize{10}{10}\selectfont
    \caption{Human Labeling Error}
   \label{tab:human}
        \centering 
   \begin{tabular}{l | l | l}
      \hline
      Metric & Median & Mean\\
      \hline 
      $E_R$ (deg) & 1.71 & 2.11\\
      $E_T$ (meters) & 1.91 & 3.66\\
      $E_{TN}$ & 0.011 & 0.020\\
      $E_C$ & 0.045 & 0.056\\
   \end{tabular}
\end{table}

The images are hand-labeled with pose information using a custom-built tool\footnote{\url{https://github.com/autognc/opat-js}} that overlays a 3D model of the spacecraft on top of each image and allows the user to line it up with the real spacecraft. We also estimate the human error associated with this labeling tool by creating a synthetic dataset with the same number of images and similar poses as the real image dataset. This gives access to known ground-truth poses so that the human labeling error can be measured. The results are presented in Table~\ref{tab:human}. We find that the human labeling error is small enough to still conduct a robust analysis of the pose estimation system's error.

\newpage

\begin{table}[h]
	\fontsize{10}{10}\selectfont
    \caption{Real Dataset Performance}
   \label{tab:real_results}
        \centering 
   \begin{tabular}{r | l | l | l}
      \cline{2-4} 
      & Metric & Median & Mean\\
      \hline 
      \multirow{2}{*}{Object Detection Metrics}
      & RoI Accuracy & - & 0.95\\
      & IoU & 0.90 & 0.85\\
      \hline
      \multirow{4}{*}{Pose Metrics}
      & $E_R$ (deg) & 7.33 & 35.26\\
      & $E_T$ (meters) & 7.36 & 112.83\\
      & $E_{TN}$ & 0.032 & 0.317\\
      & $E_C$ & 0.170 & 0.932\\
      \hline
      \multirow{5}{*}{\specialcell{Rejected Estimates \\ ($\hat{E_k} > 20$) Removed}}
      & Proportion Rejected & - & 0.16\\
      & $E_R$ (deg) & 6.48 & 29.88\\
      & $E_T$ (meters) & 6.19 & 10.94\\
      & $E_{TN}$ & 0.026 & 0.040\\
      & $E_C$ & 0.156 & 0.561\\
   \end{tabular}
\end{table}

Table~\ref{tab:real_results} shows the performance of the pose estimation system on the real images. The object detection network still performs quite well. The pose errors once again contain some ``bad estimates'' on the more difficult images. Keeping the same $\hat{E_k}$ threshold of 20 pixels, the error prediction network rejects 16\% of the estimates and easily recognizes all of the most extreme outliers. Overall, considering the relative difficulty of the real image test set, the system generalizes well.

\begin{figure}[htp]
    \centering 
\begin{subfigure}[t]{0.3\textwidth}
  \includegraphics[width=\linewidth]{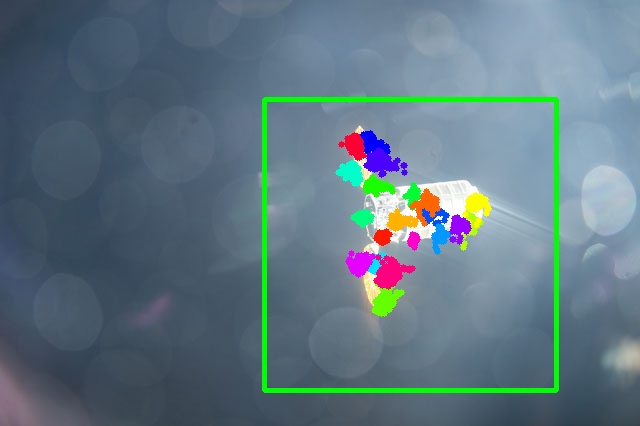}
  \caption{$E_R = 4.22\degree$, $E_{TN} = 0.013$ \\ Noisy keypoint estimates can still produce a good pose estimate by using RANSAC.}
\end{subfigure}\hfil 
\begin{subfigure}[t]{0.3\textwidth}
  \includegraphics[width=\linewidth]{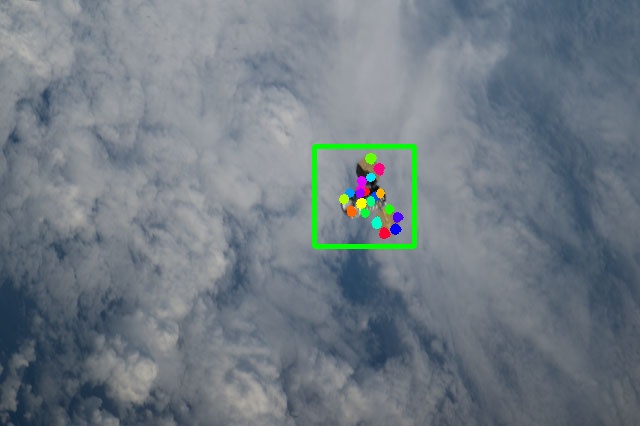}
  \caption{$E_R = 2.72\degree$, $E_{TN} = 0.034$ \\ The object detection network performs well at multiple scales.}
\end{subfigure}\hfil 
\begin{subfigure}[t]{0.3\textwidth}
  \includegraphics[width=\linewidth]{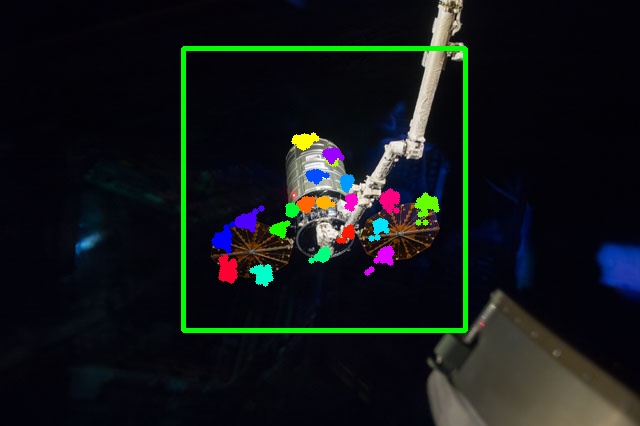}
  \caption{$E_R = 6.34\degree$, $E_{TN} = 0.045$ \\ The system performs well even with an object slightly occluding Cygnus, which does not appear at all in the training data.}
\end{subfigure}

\medskip
\begin{subfigure}[t]{0.3\textwidth}
  \includegraphics[width=\linewidth]{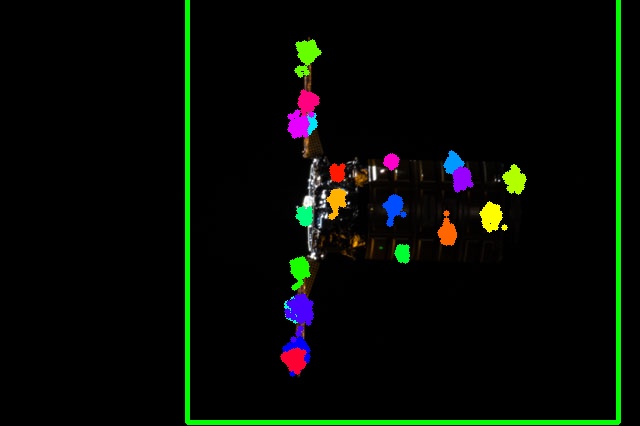}
  \caption{$E_R = 177.6\degree$, $E_{TN} = 0.025$ \\ In dark lighting, the network cannot see the logos and guesses the incorrect orientation around the barrel.}
\end{subfigure}\hfil 
\begin{subfigure}[t]{0.3\textwidth}
  \includegraphics[width=\linewidth]{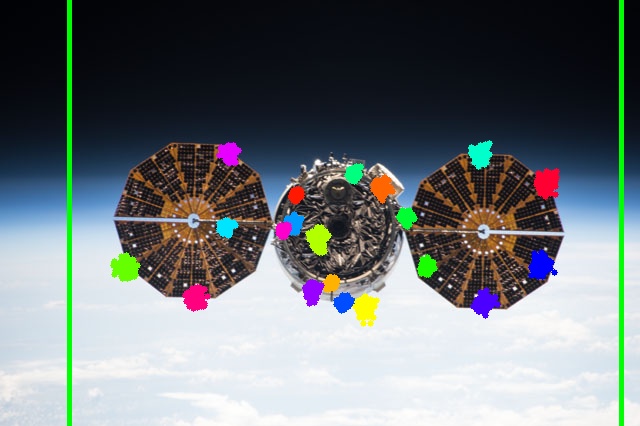}
  \caption{$E_R = 179.5\degree$, $E_{TN} = 0.009$ \\ This viewing angle obscures the logos, again making it very difficult for the network to determine the correct orientation.}
\end{subfigure}\hfil 
\begin{subfigure}[t]{0.3\textwidth}
  \includegraphics[width=\linewidth]{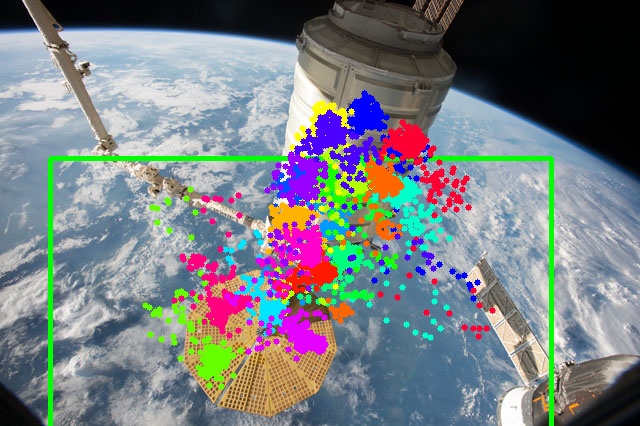}
  \caption{$E_R = 17.50\degree$, $E_{TN} = 0.105$ \\ This is an obvious failure case, which is rejected easily by the error prediction network ($\hat{E_k} = 86$).}
\end{subfigure}
\caption{The pose estimation system applied to the 6 real images from Figure~\ref{fig:real_images}. The green box is the RoI from the object detection network, and the colored dots are the output from the keypoint detection network same as in Figure~\ref{fig:keypoint_example}.}
\label{fig:real_image_detections}
\end{figure}

However, the error prediction network notably does not help very much with the mean rotation error compared to the mean translation error. This is because the Cygnus spacecraft has a major symmetry: namely, a 180\degree~rotation around the barrel is difficult to distinguish, indicated primarily by the presence or absence of two logos. This poses an especially challenging obstacle to the keypoint regression network when generalizing to real data. In some images, the network misses the subtle features necessary to determine the correct orientation, due to the fact that it was trained using an imperfect synthetic model. This particular type of mistake produces a confident estimate that is unlikely to be rejected by the error prediction network, yet has near 180\degree~rotation error (see Figure~\ref{fig:real_image_detections} (d) and (e) for examples).

\begin{table}[htp]
	\fontsize{10}{10}\selectfont
    \caption{Rotation Error Allowing Symmetrical Orientations}
   \label{tab:real_flip_results}
        \centering 
   \begin{tabular}{r | l | l | l}
      \cline{2-4} 
      & Metric & Median & Mean\\
      \hline 
      \multirow{2}{*}{Pose Metrics}
      & $E_R$ (deg) & 6.20 & 11.97\\
      & $E_C$ & 0.154 & 0.526\\
      \hline
      \multirow{2}{*}{\specialcell{Rejected Estimates \\ ($\hat{E_k} > 20$) Removed}}
      & $E_R$ (deg) & 5.38 & 8.37\\
      & $E_C$ & 0.136 & 0.186\\
   \end{tabular}
\end{table}

Table~\ref{tab:real_flip_results} presents updated rotation and combined error values if the issue of symmetry is ignored; i.e., the rotation error is computed as the smallest angle rotation that aligns the estimated pose with either of the two ambiguous orientations. These values correspond to the feasible mission scenario where the orientation of Cygnus about this axis of symmetry is not important, only the orientation of its overall shape and major features. Without the 180\degree~outliers, the pose estimation system achieves a rotation error within 10\degree~and a combined error lower than it does on the much easier SPEED real image test set.

\begin{figure}[htp]
    \centering 
\begin{subfigure}[t]{0.5\textwidth}
  \includegraphics[width=\linewidth]{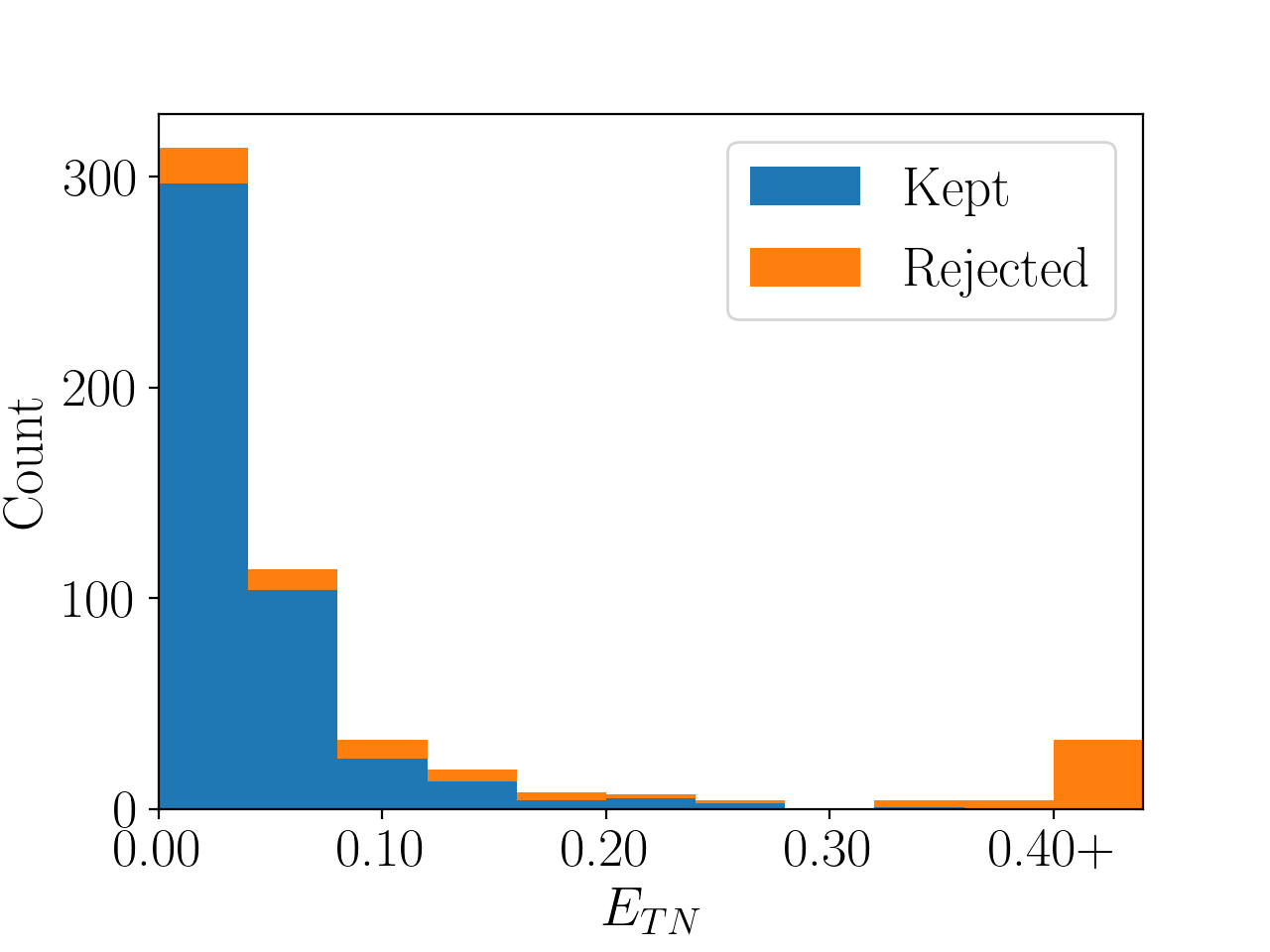}
\end{subfigure}\hfil 
\begin{subfigure}[t]{0.5\textwidth}
  \includegraphics[width=\linewidth]{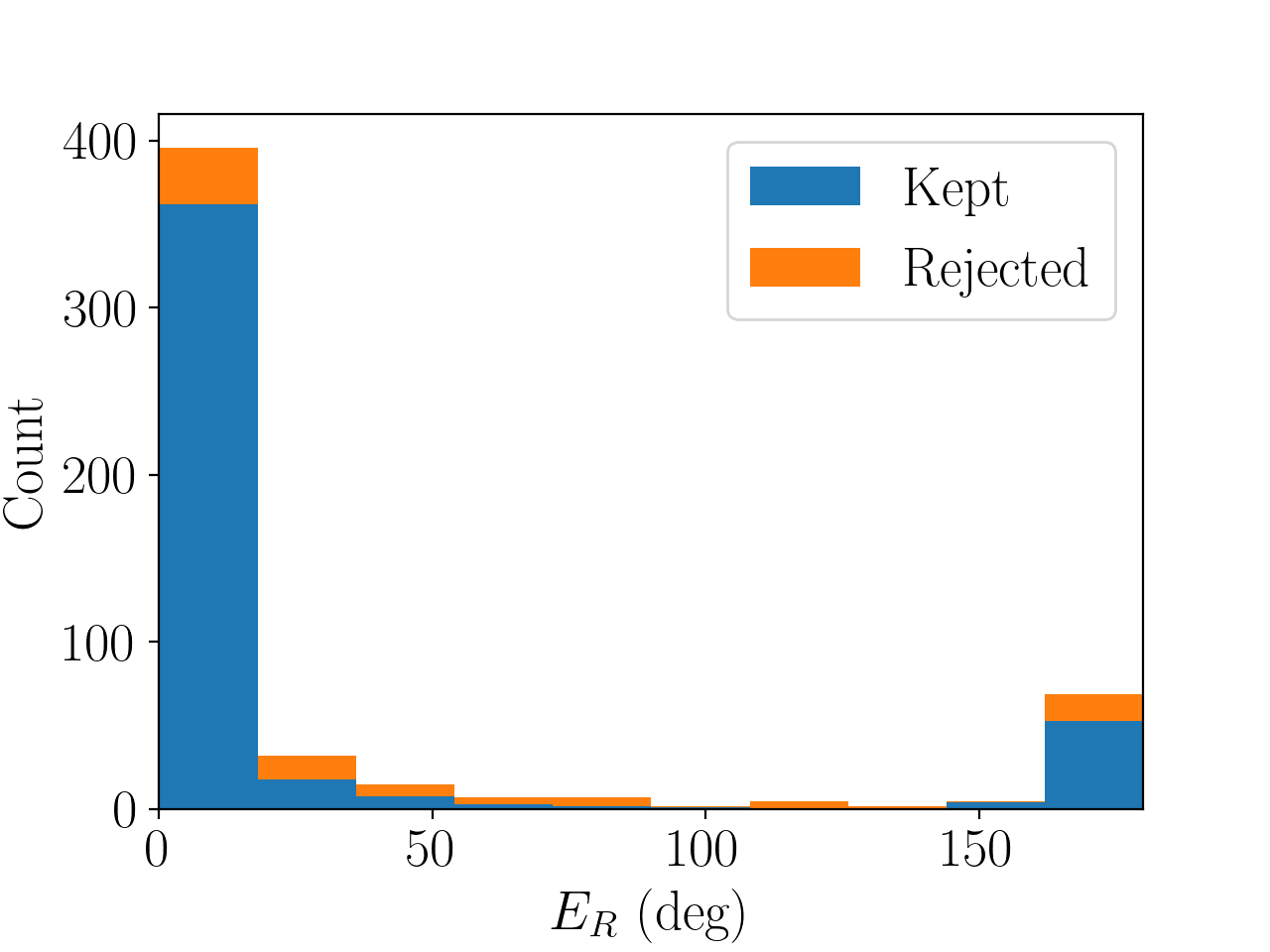}
\end{subfigure}\hfil 
\caption{Histogram of translation error (left) and rotation error (right) that also shows images rejected by the error prediction network. The error prediction step helps immensely with translation error, rejecting nearly all of the outliers. However, it is much less effective for rotation error, due to the rotational symmetry. All of the outliers near 180\degree~are otherwise ``good estimates'' with the incorrect orientation (e.g. Figure~\ref{fig:real_image_detections} (d) and (e)).}
\label{fig:hist}
\end{figure}

\subsection{Hardware Performance}
To benchmark the pose estimation model, we use the Intel Joule 570x single-board computer. It has been used previously in a CNN-based visual navigation system onboard a 3U CubeSat\cite{Seeker1}, establishing its viability as flight-ready commercial off-the-shelf (COTS) hardware. The computational capabilities of the Joule are severely limited in comparison to typical ground-based hardware such as a laptop or smartphone. Achieving real-time performance on the device demonstrates the suitability of our system for nearly any realistic mission hardware.

The Joule contains a 1.7GHz Intel Atom processor, 4 GB of RAM, and no dedicated graphics card. With these specifications, the pose estimation model runs at approximately 0.56 Hz using TensorFlow\cite{tensorflow}. However, a default TensorFlow installation is not optimized for inference on low-power devices. Proper optimization for low-power devices is known to increase speed by 10-20 times. The most common toolkit for such optimization is TensorFlow Lite\cite{tensorflow}; however, TensorFlow Lite is optimized for the ARM architecture that is dominant among mobile devices. The analogous toolkit for x86 Intel processors is OpenVINO\footnote{\url{https://software.intel.com/content/www/us/en/develop/tools/openvino-toolkit.html}}. By converting the pose estimation system into the OpenVINO format, inference speed increased to 6.6 Hz without a reduction in model accuracy.

\begin{table}[h]
	\fontsize{10}{10}\selectfont
    \caption{Inference Speed Comparison (ms)}
   \label{tab:inference_speed}
        \centering 
   \begin{tabular}{ c | c | c | c | c | c }
      \hline 
      Tool & Object Detection & Keypoint Regression & PnP Solver & Error Prediction & \textbf{Combined} \\
      \hline 
      TensorFlow & 710 & 434 & 10 & 621 & \textbf{1777} \\
      OpenVINO  & 68 & 31 & 10 & 41 & \textbf{152} \\
   \end{tabular}
\end{table}

Table~\ref{tab:inference_speed} presents inference times for each component of the system. Timing data was collected by measuring the average inference time over over a period of 10 minutes, allowing the Joule reach thermal equilibrium. Once at thermal equilibrium, the power consumption for all models averaged 3.7 Watts.

\section{Conclusion}

The presented pose estimation system achieves state-of-the-art accuracy on the Spacecraft Pose Estimation Dataset (SPEED). At the same time, it runs at 6.6 Hz --- fast enough to be considered real-time for most applications --- on low-power commercial-off-the-shelf (COTS) hardware suitable for small satellites. Most importantly, given a complete lack of real training data, the system is still able to perform accurately on real images with difficult and diverse conditions. In the case of particularly difficult images or random failures, the error prediction network is able to automatically filter out bad estimates. 

As such, the presented work comprises a nearly flight-ready pose estimation system. It is suitable to run in real-time on a small satellite with limited power requirements and a single monocular camera. The synthetic image generation and training techniques are easily applicable to any spacecraft. Overall, the system's real image performance is good enough to be useful for many applications in autonomous proximity operations; examples include formation flying, debris removal, and on-orbit inspection or servicing.

Future work will focus on integrating the pose estimation system with the other components of a fully flight-ready system. Primarily, this includes applying a dynamics-aware filtering algorithm (such as a Kalman filter) to the produced pose estimates. However, properly evaluating such a fully integrated system requires labeled time-series data from real proximity operations, which is even more difficult to obtain than ordinary real images. Another avenue of future work is to further address the issue of rotational symmetry, which is the biggest weakness of the presented pose estimation system. There exists prior work that aims to solve the specific problem of rotational symmetry in pose estimation, which could possibly be applied to this work.

\section{Acknowledgements}
We would like to thank Siddarth Kaki for his support and feedback on a variety of research topics. We also thank Evan Wilde for his aid with Blender modeling. Finally, we thank the NASA Johnson Space Center, in particular Sam Pedrotty, which provided funding that contributed to this work.



\bibliographystyle{AAS_publication}   
\bibliography{references}   

\end{document}